\documentclass[11pt]{article}
\pdfoutput=1

\usepackage[preprint]{acl}

\usepackage{times}
\usepackage{latexsym}

\usepackage[T1]{fontenc}

\usepackage[utf8]{inputenc}

\usepackage{microtype}

\usepackage{inconsolata}

\usepackage{graphicx}

\usepackage{amsmath}
\usepackage{amssymb}
\usepackage{booktabs}
\usepackage{enumitem}
\usepackage{xspace}
\usepackage{tikz}
\usetikzlibrary{positioning, arrows.meta, calc, fit, backgrounds}

\newcommand{\dataset}{\textsc{WorkflowPerturb}\xspace}
\newcommand{\Dataset}{\textsc{WorkflowPerturb}\xspace}

\newcommand{\censor}[1]{xxxx}

\definecolor{WowColor}{rgb}{.75,0,.75}
\definecolor{SubtleColor}{rgb}{0,0,.50}

\newcounter{margincounter}

\renewcommand{\cite}[1]{\citep{#1}}

\title{\dataset: Calibrated Stress Tests for Evaluating Multi-Agent Workflow Metrics}

\author{
  \textbf{Madhav Kanda\textsuperscript{1}},
  \textbf{Sharad Agarwal\textsuperscript{2}},
  \textbf{Rodrigo Fonseca\textsuperscript{2}},
  \textbf{Alok Gautam Kumbhare\textsuperscript{2}},
  \textbf{Pedro Las-Casas\textsuperscript{2}}
\\
\\
  \textsuperscript{1}University of Illinois Urbana-Champaign \quad
  \textsuperscript{2}Microsoft
}

\begin{document}
\maketitle
\begin{abstract}
Multi-agent LLM systems that generate structured workflows from
natural-language requests are now deployed in production across cloud
automation, DevOps, and enterprise orchestration. Operating them
exposes a recurring change-management problem. Routine updates, such as
re-running an input, swapping the LLM, or refactoring an
agent's prompt or orchestration code, often produce workflows that differ
substantially from validated references. Engineers then lack a principled
way to decide whether a change is safe to ship.
Automatic workflow evaluation is the natural tool, but in practice
metric scores are poorly calibrated, and a
numeric change rarely communicates the severity of the underlying
degradation. We introduce \dataset{}, a controlled benchmark that applies
realistic, graded perturbations to
golden workflows. It contains 4{,}973 golden workflows and 44{,}757
perturbed variants across three perturbation types (Missing Steps, Compressed
Steps, Description Changes) at severities of 10\%,
30\%, and 50\%. We benchmark multiple metric families, analyzing their
sensitivity and calibration using expected score trajectories and
instance-level alert rates.
Our results characterize systematic differences across families and
support severity-aware interpretation of workflow evaluation scores in
change-management settings. \dataset{} is publicly
available.\footnote{\url{https://huggingface.co/datasets/microsoft/WorkflowPerturb}}
\end{abstract}

\section{Introduction}

Large Language Models (LLMs) are increasingly used to generate multi-step,
dependency-rich workflows for high-stakes settings, where failures can have real
operational consequences. Such workflows arise in domains
including scientific automation, biomedical information access, question
answering, enterprise copilots, and cloud and DevOps systems
\cite{bran2023chemcrowaugmentinglargelanguagemodels, jin2024genegpt,
microsoft_powerautomate_copilot, aws_amazonq_cli,
dalal-etal-2026-compositional}. In these settings, correctness depends not only
on surface-level textual similarity, but on the presence of required
steps and the correctness of their ordering and dependencies.

Such workflows are produced by multi-agent systems deployed in production
\cite{microsoft_powerautomate_copilot, aws_amazonq_cli}. This work grew out of
operating one of them, \textsc{LLexus}~\cite{lascasas2024llexus}, for over two
years. \textsc{LLexus} compiles a natural-language troubleshooting guide into an
executable incident-response workflow: a cascade of LLM agents selects tools,
extracts the workflow structure, fills in each step, and validates the result,
yielding a DAG that mixes deterministic tool invocations with narrowly scoped
semantic steps. Since the expensive reasoning happens once at authoring time
rather than at every incident, each workflow is reviewed by an engineer,
versioned alongside its guide, and executed repeatedly against live incidents.
That review gate is where workflow evaluation must operate: every regenerated
workflow must be judged against an already-approved reference.

Routine system changes frequently produce workflows that differ
substantially from those references. Three classes of change are
particularly disruptive: (i)~re-running the same request under stochastic
decoding yields a structurally different workflow; (ii)~swapping in a newer
LLM (sometimes a forced upgrade when older models are
deprecated~\citep{azure_openai_model_retirements}) shifts the distribution of
generated steps; and
(iii)~refactoring an agent's prompts, tool inventories, or orchestration code
alters which steps are emitted and how they are sequenced. Each forces
engineers to decide, before shipping, whether the new workflow is functionally
equivalent to its reference or is a silent regression that could cause an outage.
Manual review does not scale. Automatic workflow metrics are the natural
alternative, but they proved poorly calibrated in practice, with numeric changes
rarely communicating the \emph{severity} of degradation. The benchmark presented
here abstracts the degradation patterns we observed in \textsc{LLexus} into
controlled perturbations that apply to any workflow corpus, so the study is
reproducible without access to that system.

Evaluating these workflows remains unexpectedly difficult. Existing workflow
metrics each capture only a narrow slice of correctness: lexical metrics fail
under paraphrasing, structural metrics may overlook semantic drift, and semantic
metrics can assign favorable scores to workflows missing critical steps. A
central practical challenge is calibration: it is often unclear what a score
means in terms of functional risk. For example, does a drop from 0.90 to 0.84
reflect harmless rephrasing or the loss of an essential dependency? Without
calibration, metric values are hard to use for regression
testing, model comparison, or automated filtering.

The consequences of getting this wrong are severe. Flawed orchestration can
cause outages or misconfigured infrastructure, and subtle workflow deviations
can alter analytical conclusions in scientific and biomedical domains. This
sharpens the practical question: how can we determine whether a
modified workflow remains functionally equivalent to a validated reference, and
integrate such checks into CI/CD without manual review?

We introduce \dataset{}, a benchmark of over
44{,}000 workflows constructed through controlled perturbations. We generate
three perturbation types, Missing Steps, Compressed Steps, and Description
Changes, each applied at graded intensity levels of 10\%, 30\%, and 50\%. These
structured degradations enable systematic analysis of metric calibration under
known severity, revealing which metrics are robust to benign edits, which fail
under functional degradation, and how to interpret workflow scores reliably.

\section{Related Work}

\paragraph{Agent and workflow evaluation.}
Recent benchmarks evaluate LLMs in agentic and tool-using
settings, where models must select, invoke, and sequence external functions:
e.g., the Berkeley Function Calling Leaderboard~\cite{patilberkeley} emphasizes
end-to-end task success in tool-augmented environments. A complementary line of
work analyzes execution trajectories to diagnose and localize agent failures,
including AGENTRX~\cite{barke2026agentrxdiagnosingaiagent} (intermediate
actions, tool calls, and environment feedback) and
TRAIL~\cite{deshpande2025trailtracereasoningagentic} (trace-based reasoning over
structured action sequences). Both families operate on runtime traces and
assume access to environment interactions or success signals; in contrast, we
study static workflow representations and evaluate how automatic metrics
respond when workflows deviate from golden references in controlled and graded
ways, enabling systematic analysis of metric calibration, sensitivity, and
blind spots. Adaptive execution frameworks such as
VineLM~\cite{pagonas2026vinelm} select a different LLM at each
configurable workflow stage under per-request cost and latency budgets. They
further expand the space of outputs a single workflow can produce across runs,
sharpening the need for calibrated cross-run metric comparisons of the kind we
study here.

\paragraph{LLM-as-a-Judge and Its Limitations.}
LLMs have emerged as scalable automatic evaluators with
strong agreement to human judgments
\citep{zheng2023judgingllmasajudgemtbenchchatbot,
chiang2023largelanguagemodelsalternative}, but recent work highlights important
reliability failures.
\citet{shi-etal-2025-judging} systematically study position bias across 15
judges and 22 tasks and find that judgments depend strongly on candidate
ordering and on the quality gap between candidates. These are properties of how
options are \emph{presented} to the judge, rather than of the underlying
artifact. Related work also documents biases such as length, self-preference,
and stylistic effects \citep{wang2023largelanguagemodelsfair}. These studies
characterize fragility along one axis: \emph{perturbing the inputs shown to
the judge}. Our perspective is complementary and orthogonal: we hold the
judge's prompt and presentation fixed and instead perturb the
\emph{artifact under judgment} at calibrated severities, measuring how a
judgment-based metric responds to known functional degradation.

\section{\dataset{}}

\begin{figure*}[t!]
    \centering
    \tikzset{
        wnode/.style={draw, rounded corners=2pt, inner sep=2pt, align=center,
                      font=\scriptsize, minimum width=1.7cm, minimum height=0.6cm,
                      line width=0.4pt},
        wblue/.style={wnode, fill=blue!18},
        wred/.style={wnode, fill=red!22},
        warrow/.style={-{Latex[length=1.3mm,width=1.1mm]}, line width=0.4pt},
    }
    \begin{minipage}{0.95\textwidth}
        \centering
        \footnotesize\itshape
        Task: handle a customer-support escalation for a faulty smart-home device.
    \end{minipage}
    \par\vspace{3pt}
    \begin{minipage}[t]{0.245\textwidth}\centering
        \textbf{\small Original Workflow}\par\vspace{2pt}
        \begin{tikzpicture}[node distance=0.28cm]
            \node[wnode]                                          (a1) {Retrieve Customer\\ticket};
            \node[wnode, below=0.45cm of a1, xshift=-0.95cm]      (a2) {Check Device\\Diagnostic};
            \node[wnode, below=0.45cm of a1, xshift= 0.95cm]      (a3) {Search known\\issue database};
            \node[wblue, below=1.7cm of a1]                       (a4) {Draft Troubleshooting\\steps};
            \node[wred,  below=of a4]                             (a5) {Generate Summary\\Report};
            \node[wnode, below=of a5]                             (a6) {Notify customer\\\& update CRM};
            \draw[warrow] (a1.south) -- (a2.north);
            \draw[warrow] (a1.south) -- (a3.north);
            \draw[warrow] (a2.south) -- (a4.north);
            \draw[warrow] (a3.south) -- (a4.north);
            \draw[warrow] (a4) -- (a5);
            \draw[warrow] (a5) -- (a6);
        \end{tikzpicture}
    \end{minipage}\hfill
    \begin{minipage}[t]{0.245\textwidth}\centering
        \textbf{\small Missing Steps}\par\vspace{2pt}
        \begin{tikzpicture}[node distance=0.28cm]
            \node[wnode]                                          (b1) {Retrieve Customer\\ticket};
            \node[wnode, below=0.45cm of b1, xshift=-0.95cm]      (b2) {Check Device\\Diagnostic};
            \node[wnode, below=0.45cm of b1, xshift= 0.95cm]      (b3) {Search known\\issue database};
            \node[wblue, below=1.7cm of b1]                       (b4) {Draft Troubleshooting\\steps};
            \node[wnode, below=of b4]                             (b6) {Notify customer\\\& update CRM};
            \draw[warrow] (b1.south) -- (b2.north);
            \draw[warrow] (b1.south) -- (b3.north);
            \draw[warrow] (b2.south) -- (b4.north);
            \draw[warrow] (b3.south) -- (b4.north);
            \draw[warrow] (b4) -- (b6);
        \end{tikzpicture}
    \end{minipage}\hfill
    \begin{minipage}[t]{0.245\textwidth}\centering
        \textbf{\small Compressed Steps}\par\vspace{2pt}
        \begin{tikzpicture}[node distance=0.28cm]
            \node[wnode]                                          (c1) {Retrieve Customer\\ticket};
            \node[wnode, below=0.45cm of c1, xshift=-0.95cm]      (c2) {Check Device\\Diagnostic};
            \node[wnode, below=0.45cm of c1, xshift= 0.95cm]      (c3) {Search known\\issue database};
            \node[wblue, below=1.7cm of c1]                       (c4) {Draft Troubleshooting\\\& summary report};
            \node[wnode, below=of c4]                             (c6) {Notify customer\\\& update CRM};
            \draw[warrow] (c1.south) -- (c2.north);
            \draw[warrow] (c1.south) -- (c3.north);
            \draw[warrow] (c2.south) -- (c4.north);
            \draw[warrow] (c3.south) -- (c4.north);
            \draw[warrow] (c4) -- (c6);
        \end{tikzpicture}
    \end{minipage}\hfill
    \begin{minipage}[t]{0.245\textwidth}\centering
        \textbf{\small Description Changes}\par\vspace{2pt}
        \begin{tikzpicture}[node distance=0.28cm]
            \node[wnode]                                          (d1) {Access Customer\\ticket};
            \node[wnode, below=0.45cm of d1, xshift=-0.95cm]      (d2) {Review Device\\Diagnostics};
            \node[wnode, below=0.45cm of d1, xshift= 0.95cm]      (d3) {Consult known\\issue database};
            \node[wblue, below=1.7cm of d1]                       (d4) {Prepare troubleshooting\\steps};
            \node[wred,  below=of d4]                             (d5) {Make Summary\\Report};
            \node[wnode, below=of d5]                             (d6) {Revise CRM \&\\notify the customer};
            \draw[warrow] (d1.south) -- (d2.north);
            \draw[warrow] (d1.south) -- (d3.north);
            \draw[warrow] (d2.south) -- (d4.north);
            \draw[warrow] (d3.south) -- (d4.north);
            \draw[warrow] (d4) -- (d5);
            \draw[warrow] (d5) -- (d6);
        \end{tikzpicture}
    \end{minipage}
    \caption{Real-world example illustrating perturbations in LLM-generated
        workflows for a customer support escalation task.
        \textsc{Missing Steps:} The \textcolor{red!75!black}{red node} present
        in the original workflow is omitted. \textsc{Compressed Steps:} The
        \textcolor{red!75!black}{red} and \textcolor{blue!75!black}{blue}
        nodes are merged into a single \textcolor{blue!75!black}{blue} node,
        reducing granularity. \textsc{Description Changes:} All nodes retain the
        same semantic meaning as the original but differ in syntactic phrasing.}
    \label{fig:workflow}
    \vspace{-.05in}
\end{figure*}

We model a workflow as a directed acyclic graph (DAG) $G=(V,E)$, where each node
$v \in V$ represents a step described in natural language and each edge $(u,v)
\in E$ denotes a precedence constraint. Given a golden workflow $G$ and a
candidate workflow $G'$, an evaluation metric produces a score $s(G,G') \in
[0,1]$ (or a normalized variant) intended to reflect workflow quality or
similarity. Rather than proposing a new workflow generator, our focus is to
analyze \emph{metric behavior}: under controlled degradations of known severity,
do metric scores change in an interpretable and calibrated manner? We study this
question by introducing structured perturbation types, graded severity levels,
and an expected score trajectory characterizing ideal metric response under
degradation.

\subsection{Benchmark Construction}
\label{sec:benchmark-construction}

\noindent{\textbf{Source Data.}} We begin with the workflows in the training
split of WORFBENCH~\cite{qiao2024benchmarking} and retain those with \(\lvert V \rvert
\ge 5\) nodes (\textbf{at least} five steps). This lower threshold ensures
each workflow has enough steps to meaningfully perturb and evaluate metric
sensitivity. We impose no upper cap, so the corpus spans
short-to-long workflows: 5--6--41 nodes (min--median--max), with 13
workflows containing 20 or more nodes. The resulting corpus contains
$4{,}973$ golden workflows drawn from seven source datasets (WikiHow,
ALFWorld, WebShop, Lumos, ToolBench, OS, and ToolAlpaca; WikiHow and
ALFWorld together account for $86\%$ of the corpus). Combined with three
perturbation types at three severity levels, this yields $44{,}757$
perturbed variants in the released benchmark.\\

\noindent{\textbf{Perturbation Patterns.}} From more than two years of operating
\textsc{LLexus}~\cite{lascasas2024llexus} in production, three recurring degradation patterns
emerged. We model them as controlled perturbations applied to golden workflows,
illustrated on a customer-support escalation task in Figure~\ref{fig:workflow}:

\begin{itemize} [leftmargin=*, itemsep=2pt, parsep=0pt, topsep=2pt]

 \item \textbf{Missing Steps:} LLM-generated workflows often omit essential
actions that are required for complete task execution. Such omissions result in
incomplete or truncated execution paths when compared to the golden workflow. 

 \item \textbf{Compressed Steps:} Generated workflows frequently contain fewer
nodes than their golden counterparts, condensing multiple fine-grained actions
into a single broader step. While this may preserve most of the content, it
alters the structural properties of the workflow and leads to mismatches under
graph-based evaluation. 

  \item \textbf{Description Changes:} Even when workflows preserve the correct
  structure, step descriptions often differ syntactically from the golden
  reference. This lexical variation confuses surface metrics while preserving
  semantics.

\end{itemize}

\noindent\textbf{Diagnostic mapping.} These three patterns are not redundant
along the metric axis (the seven metrics named below are formally defined in
Section~\ref{sec:metrics}). \textsc{Missing Steps} most distinctly stresses
structural (Graph/Chain F1) and judgment-based metrics by reducing required
nodes; \textsc{Compressed Steps} stresses both structural and ordering metrics
(Kendall's $\tau$) by collapsing edges and reshuffling precedence; and
\textsc{Description Changes} most directly stresses lexical metrics (BLEU, GLEU)
while leaving the ground-truth topology unchanged: the node set and edge set are
preserved, and only the node descriptions are paraphrased. A preserved topology,
however, does not make the \emph{structural metrics} invariant. Graph and Chain
F1 align nodes semantically using Sentence-BERT at $\tau_{\text{align}}=0.8$
(Appendix~\ref{appendix:metrics}), so heavy paraphrasing can push a still-present
node below the alignment threshold and lower the score even though no node or
edge was changed. Empirically, Graph F1 falls below $1.0$ for $19.55\%$,
$42.47\%$, and $59.18\%$ of \textsc{Description Changes} variants at the $10\%$,
$30\%$, and $50\%$ levels, reaching as low as $0.50$, $0.38$, and $0.27$
respectively. Each perturbation type is therefore a useful, though not perfectly
isolated, diagnostic for a different metric family
(see Section~\ref{sec:quant_results}).

To study the impact of these issues on evaluation metrics, we apply each
perturbation type at three severity levels \(\{10\%, 30\%, 50\%\}\) of the golden
workflow (an illustration for \textsc{Missing Steps} appears as
Figure~\ref{fig:diff-percent-perturb} in the appendix). Throughout, \emph{severity}
denotes this \emph{perturbation fraction}, the share of nodes edited, rather than
perceived or execution-level risk; it gives a controlled, reproducible ordering
against which to probe whether scores degrade proportionally.\\

\noindent{\textbf{Perturbation Generation and Validation.}} For each workflow
and perturbation configuration, we employ LLMs (GPT-4o as the primary
generator, with GPT-4.1 and o3-mini as fallbacks for variants that
initially fail the static checks below) to generate variants. To ensure
correctness, each variant undergoes automated static validation checks:

\begin{itemize} [leftmargin=*, itemsep=2pt, parsep=0pt, topsep=2pt]
  \item \textbf{Node Count Consistency:} Given a golden workflow with 10 nodes,
  applying a 10\% removal perturbation must produce a workflow with precisely 9
  nodes. Any deviation from this expected count is flagged as inconsistent.
  \item \textbf{Change Count Verification:} For a $30\%$ perturbation on a
  10-node workflow, exactly $3$ nodes must be altered (removed, compressed, or
  paraphrased, depending on type). If fewer or more are modified, the output is
  rejected. For removal and compression, verification compares node counts
  before and after. For paraphrasing under the primary path, verification counts
  how many step descriptions changed; the LLM chooses which $N$ nodes to
  paraphrase under that constraint.
\end{itemize}

These checks target the two most common generation failure modes: (i) deleting
or merging too many nodes, and (ii) over- or under-applying paraphrasing. Each
generator call prepends a fixed few-shot example (one per perturbation type)
before the row-specific instruction: the golden workflow plus the perturbation
directive. Variants that fail validation enter a feedback-driven refinement
loop, escalating to stronger models on repeated failure. The loop is effective:
the initial GPT-4o pass left $17.7\%$, $27.5\%$, and $44.7\%$ of Missing-Steps,
Compressed-Steps, and Description-Changes variants failing the static checks, and
successive fallbacks reduced these to $0.03\%$, $0.14\%$, and $1.3\%$ ($4$, $21$,
and $200$ generation-time residual variants, respectively). A separate recount
of the final dataset flagged $175$ node-count mismatches, distinct from these
residuals. We also manually inspected $90$ variants ($10$ per perturbation type
and severity) for perturbation compliance, of which $89/90$ ($98.9\%$) passed,
with row-level decisions shipped alongside the dataset. Prompt templates,
fallback handling, the node-count audit, the inspection protocol, and the
description-change pipeline appear in Appendix~\ref{appendix:prompts}. \\

\noindent\textbf{Score Assignment and Perturbation Alignment.} Each workflow is
assigned a score reflecting perturbation severity, with lower scores indicating
deviation from the golden workflow. For \emph{Missing Steps} and
\emph{Compressed Steps}, the score equals the remaining workflow fraction,
computed as $1 - p$ for perturbation percentage $p$ (e.g., $p{=}30\% \rightarrow
0.7$, $p{=}50\% \rightarrow 0.5$). In the case of Compressed Steps, this
decrease is justified by the loss of structural granularity: multiple
fine-grained actions (often corresponding to distinct tool invocations) are
merged into broader nodes, violating the one-tool-per-step abstraction and
reducing fidelity to the golden execution graph. For \emph{Description Changes},
the score remains constant across severity levels, since these edits affect only
textual descriptions and not the underlying workflow structure.

\section{Metrics}
\label{sec:metrics}

We evaluate seven metrics across five families. Structural metrics (Graph F1, Chain F1) measure
topological consistency via subgraph matching and sequence alignment. Lexical
metrics (BLEU, GLEU) compute $n$-gram overlap. BERTScore measures contextual
embedding similarity. Kendall's $\tau$ evaluates rank-order consistency.
LLM-as-Judge provides rubric-based holistic assessment.
Full definitions appear in Appendix~\ref{appendix:metrics}.

\subsection{Structural Metrics}

\noindent\textbf{Chain F1.} Following \citet{qiao2024benchmarking}, predicted
node chains are aligned with valid topological orderings of the golden workflow,
and the longest increasing subsequence (LIS) is computed. Precision and recall
are defined as $p_{\text{chain}} = l / |V_{\text{pred}}|$ and $r_{\text{chain}}
= l / |V_{\text{golden}}|$, where $l$ is the LIS length. The final score is
their harmonic mean (F1).

\noindent\textbf{Graph F1.} The predicted and golden workflow graphs are aligned
using a Maximum Common Induced Subgraph (MCIS) algorithm. If the largest shared
subgraph contains $k$ nodes, precision and recall are defined as
$p_{\text{graph}} = k / |V_{\text{pred}}|$ and $r_{\text{graph}} = k /
|V_{\text{golden}}|$. The score is their harmonic mean (F1).

\subsection{Lexical Metrics}
\noindent\textbf{BLEU}~\cite{papineni2002bleu} is a precision-oriented $n$-gram
matching metric with a brevity penalty, while
\textbf{GLEU}~\cite{wu2016google} balances $n$-gram precision and
recall and is more robust for shorter or paraphrased step descriptions.

\subsection{Semantic Metric}
\noindent\textbf{BERTScore}~\cite{zhang2020bertscoreevaluatingtextgeneration}
compares contextual embeddings of step descriptions from pretrained transformer
models and computes precision, recall, and F1 via token-level cosine similarity.
 
\subsection{Ordering Metric}
\noindent\textbf{Kendall's $\tau$}~\cite{Kendall1938} measures rank correlation
between generated and gold step orderings, defined as $\tau = (C - D) /
\bigl(\tfrac{1}{2} n(n-1)\bigr) \in [-1,1]$, where $C$ and $D$ are the numbers
of concordant and discordant pairs among $n$ aligned steps.

\subsection{LLM-as-Judge}
To complement automatic metrics, we use an \emph{LLM-as-Judge} evaluation with
GPT-4o (Azure OpenAI). The model is prompted with the task description, golden
workflow, generated workflow, and a rubric defining 0--5 scores covering
correctness, completeness, ordering, and clarity, with anchor examples.

\section{Results and Analysis}
\label{sec:quant_results}

We evaluate metric behavior across perturbation types and severities, assigning
each workflow a predefined degradation score for interpretability. All reported
\emph{LLM-as-Judge} scores are linearly normalized from the original 0--5 scale
to the [0,1] range for comparability across metrics.

\begin{figure*}[t]
\centering
\includegraphics[width=\textwidth]{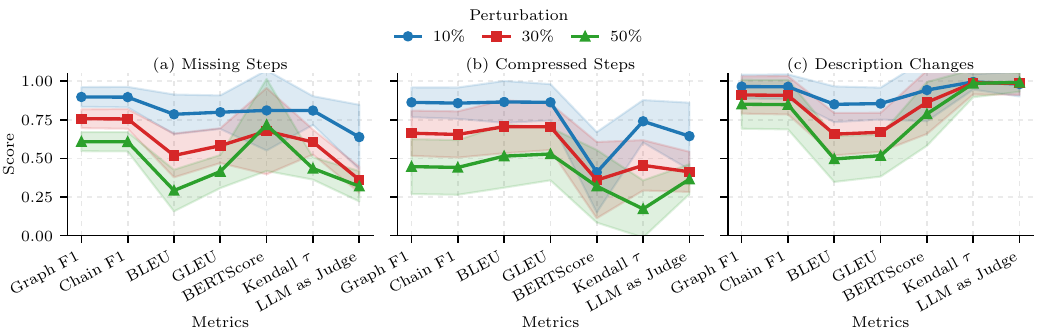}
\caption{Metric scores by perturbation type and severity (mean $\pm$ std).}
\label{fig:metrics-by-perturbation}
\end{figure*}

\paragraph{Overall trends.}
Across all perturbation types, metric scores generally decrease as severity
increases (Figure~\ref{fig:metrics-by-perturbation}), confirming that the
benchmark introduces progressively
harder deviations from the golden workflows. However, degradation patterns
differ substantially across metric families, as detailed below. Appendix
Table~\ref{tab:sensitivity-summary} condenses these into a $7\times 3$ sensitivity
matrix $\Delta_m^{\mathrm{avg}}$, defined in Section~\ref{sec:sensitivity}.

\paragraph{Missing Steps.}
Removing steps (dropped red node in Figure~\ref{fig:workflow}; numbers in
Table~\ref{tab:score_change_missing}) disrupts completeness and dependency
structure. Structural metrics degrade nearly linearly with the removal rate,
closely tracking imposed severity, while lexical metrics drop more sharply
under token loss. BERTScore is the most tolerant, and non-monotonic
(\S\ref{sec:sensitivity}). Kendall's $\tau$ and
LLM-as-Judge both decline strongly, the latter signaling missing content.

\paragraph{Compressed Steps.}
Merging fine-grained actions into broader steps (cf.~the merged red/blue nodes
in Figure~\ref{fig:workflow}; numbers in Table~\ref{tab:score_change_compress})
substantially degrades structural scores and Kendall's $\tau$, the latter
reflecting disrupted pairwise precedence under compression. Lexical metrics
decline more than under deletion, since merged descriptions break direct
$n$-gram overlap. BERTScore is already low at mild compression and falls only
slightly, suggesting embedding similarity is itself sensitive to abstraction.
LLM-as-Judge tracks the loss of procedural granularity despite preserved intent.

\paragraph{Description Changes.}
When only descriptions are modified (paraphrased-but-isomorphic graph in
Figure~\ref{fig:workflow}; numbers in Table~\ref{tab:score_change_desc}), graph
structure remains intact and structural metrics, Kendall's $\tau$, and
LLM-as-Judge stay nearly flat \emph{on average}; at the instance level, however,
structural scores are not invariant, and some paraphrase
variants fall below $1.0$ (Section~\ref{sec:benchmark-construction},
Table~\ref{tab:instance-gates}). Lexical metrics decline with heavier
paraphrasing, as expected under surface-level variation. BERTScore stays
high, unlike under structural perturbations, indicating that semantic
similarity is largely preserved.

Overall, no single metric captures all failure modes: structural and ordering
metrics detect missing or compressed steps but miss textual edits, while
lexical metrics respond strongly to paraphrasing yet over-penalize deletions;
embedding and judgment metrics are complementary.

\subsection{Sensitivity Analysis}
\label{sec:sensitivity}

\begin{figure}[t]
  \centering
  \includegraphics[width=\columnwidth]{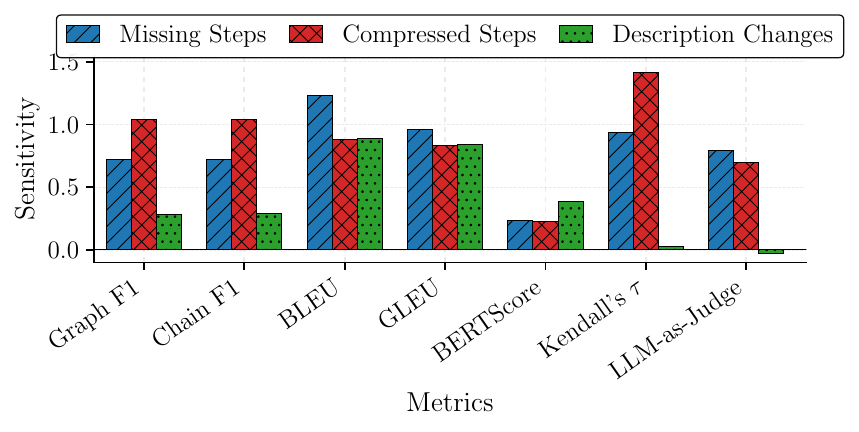}
  \caption{Per-metric sensitivity $\Delta_m^{\mathrm{avg}}$ across the three
  perturbation types.}
  \label{fig:sensitivity-analysis}
\end{figure}

To quantify how rapidly each metric degrades with severity, we summarize the
per-severity score curves by a single
\textit{sensitivity}~$\Delta_m^{\mathrm{avg}}$:
\begin{equation}
\Delta_m^{\text{avg}}
= \tfrac{1}{2}\!\left[
\tfrac{\bar{s}_m(10\%) - \bar{s}_m(30\%)}{0.20}
+ \tfrac{\bar{s}_m(30\%) - \bar{s}_m(50\%)}{0.20}
\right]
\end{equation}
where $\bar{s}_m(\alpha)$ is the mean score of metric~$m$ at $\alpha\%$
perturbation (appendix Tables~\ref{tab:score_change_missing}--\ref{tab:score_change_desc}).
Since both intervals share the $30\%$ term, this average telescopes to
$\Delta_m^{\mathrm{avg}}{=}2.5\,[\bar{s}_m(10\%){-}\bar{s}_m(50\%)]$: the $30\%$
midpoint cancels, so $\Delta_m^{\mathrm{avg}}$ alone cannot separate steady decay
from a sharp drop that then plateaus. We therefore also report the two interval
slopes, $\Delta_m^{1}$ (mild$\to$moderate) and $\Delta_m^{2}$
(moderate$\to$severe), in Appendix Table~\ref{tab:two-slope}; e.g., BERTScore under
\emph{missing steps} is non-monotonic ($\Delta^{1}{=}{+}0.67$,
$\Delta^{2}{=}{-}0.20$) behind a bland $\Delta^{\mathrm{avg}}{=}0.24$.
The high-level message: metrics whose two slopes are near-equal (the structural,
lexical, and ordering families) degrade proportionally with severity---the
behavior severity-aware gating relies on---whereas front-loaded or sign-flipping
slopes (here LLM-as-Judge and BERTScore) signal saturation or instability that a
single $\Delta^{\mathrm{avg}}$ conceals.
Figure~\ref{fig:sensitivity-analysis} visualizes $\Delta_m^{\mathrm{avg}}$;
Appendix Table~\ref{tab:sensitivity-summary} gives the numbers.

\paragraph{Structural metrics (Graph/Chain~F1).}
Peak under \emph{compressed steps} ($\Delta=1.04$); moderate for
\emph{missing steps} ($0.72$); low for \emph{description changes} ($0.29$). Both
metrics primarily track topological distortion (cf.~Figure~\ref{fig:workflow}).

\paragraph{Lexical metrics (BLEU, GLEU).}
High across all perturbation types ($\Delta\!\approx\!0.84$--$1.23$), peaking
for \emph{missing steps} where token loss dominates. They cannot, however,
distinguish structural loss from benign rewording on their own.

\paragraph{Embedding metric (BERTScore).}
Least sensitive overall ($\Delta\!<\!0.4$ everywhere): tolerant of structural
change ($0.24$ for \emph{missing steps} and $0.23$ for \emph{compressed steps})
and only modestly responsive to \emph{description changes} ($0.39$). Its
\emph{missing steps} curve is non-monotonic ($0.81\!\to\!0.68\!\to\!0.72$):
BERTScore compares only the first $\min(|V^{g}|,|V^{p}|)$ steps
(Appendix~\ref{appendix:metrics}), so deeper deletion shrinks that window
(mean $5.5\!\to\!4.1\!\to\!3.0$ steps) onto the earliest, least-displaced steps,
which score higher (at $50\%$: $0.78$ for windows of $1$--$2$ steps vs.\ $0.58$
for $6{+}$). No recommendation depends on this, since BERTScore is not in the
\textsc{Missing Steps} bundle (Table~\ref{tab:bundle-recommendation}).

\paragraph{Order-based metric (Kendall's~$\tau$).}
Strongest for \emph{compressed steps} ($1.42$); high for \emph{missing steps}
($0.94$); essentially flat for \emph{description changes} ($0.02$), consistent
with unchanged topology.

\paragraph{Judgment-based metric (LLM-as-Judge).}
Tracks structural/order behavior: high for \emph{missing steps} ($0.80$) and
moderate for \emph{compressed steps} ($0.70$); near zero, and slightly
negative within sample noise, for \emph{description changes} ($-0.03$). Its
decline is strongly front-loaded ($\Delta^{1}{=}1.38, \Delta^{2}{=}0.21$ for
missing; $1.16, 0.24$ for compressed), so as the \textsc{Missing Steps} secondary
metric it barely separates $30\%$ from $50\%$ degradation.

\paragraph{Sensitivity by workflow size.}
Since the corpus is dominated by short workflows, we recompute
$\Delta_m^{\mathrm{avg}}$ within three size strata ($5$--$7$, $8$--$15$, and
$16{+}$ nodes; $n{=}3842/1109/22$ golden workflows); Appendix
Table~\ref{tab:size-stratified} gives the full matrix. Structural sensitivity is
stable across strata ($0.73/0.70/0.69$ for Graph F1 under \emph{missing steps}),
indicating the signal is not an artifact of small graphs, and the lexical family
is likewise steady. Several metrics do diverge in the $16{+}$ stratum:
LLM-as-Judge falls to $0.32$ under \emph{missing steps} and Kendall's $\tau$ to
$0.95$ under \emph{compressed steps}. With only $22$ workflows those estimates
are unreliable, so we treat the long-workflow regime as untested rather than
validated (see Limitations).

Per-severity score tables appear in Appendix~\ref{appendix:results}.

\section{Key Insights and Practical Guidance}

\paragraph{Workflow validation under change.}
Deployed workflows evolve with model upgrades, prompt revisions, and
orchestration changes, and metric families respond differently to
structural versus textual edits (Section~\ref{sec:quant_results}). \dataset{}
enables severity-aware calibration: post-change candidates are compared
against approved goldens using a calibrated bundle
(Table~\ref{tab:bundle-recommendation}), whose assignments and thresholds
derive from the sensitivity matrix and per-severity distributions in
Appendix~\ref{appendix:results}. For \textsc{Description Changes}, however,
structural metrics are not invariant under paraphrasing
(Section~\ref{sec:benchmark-construction}), so that row's assignment is an
operating-point recommendation rather than a consequence of clean metric
specificity. For \textsc{Compressed Steps}, Kendall's $\tau$ spans a wide $0.285$
gap between its $10\%$ and $30\%$ means ($0.74\!\to\!0.46$), so the $0.65$ trip
point is set conservatively toward the mild end.
Since no single metric suffices, breaches can
trigger review or rollback, prioritizing structural regressions.

\begin{table}[t]
  \centering
  \small
  \setlength{\tabcolsep}{3pt}
  \renewcommand{\arraystretch}{1.05}
  \caption{Recommended metric bundle and alert thresholds per failure mode.}
  \label{tab:bundle-recommendation}
  \resizebox{\columnwidth}{!}{%
  \begin{tabular}{@{}lll@{\,}c@{}}
    \toprule
    \textbf{Failure mode} & \textbf{Primary} & \textbf{Secondary} & \textbf{Alert} \\
    \midrule
    Missing Steps     & Graph/Chain F1   & LLM-Judge        & $<0.75$ \\
    Compressed Steps  & Kendall's $\tau$ & Graph/Chain F1   & $<0.65$ \\
    Description Changes & BLEU/GLEU      & BERTScore        & $<0.70$ \\
    \bottomrule
  \end{tabular}%
  }
\end{table}

\paragraph{Instance-level gate behavior.}
Table~\ref{tab:bundle-recommendation}'s thresholds are read off \emph{mean}
score trajectories, so a threshold at a severity's mean fires on only about half
the individual variants there. At the instance level the recommended gates
therefore trade misses against false alarms: the \textsc{Missing Steps} gate
leaves $41.6\%$ of genuine $30\%$ deletions unflagged, and false alarms reach
$21.0\%$ on mild ($10\%$) compressions and $21.5\%$ on benign paraphrases (target
score $1.0$; the SBERT-alignment effect of
Section~\ref{sec:benchmark-construction}). We report the full per-gate firing
matrix and its derivation in Appendix~\ref{appendix:instance-gates}. These rates
are the bundle's \emph{operating characteristics}, not a defect in the metric
choices: they make the recall/false-alarm trade-off explicit so a deployment can
pick its own point rather than inherit the corpus mean. Three levers set that
point. First, moving a trip point along the score distribution trades the two
error types: a team facing costly outages raises it toward the mild-severity
mean, while one facing alert fatigue lowers it. Second, requiring the Primary
\emph{and} Secondary metric to breach jointly suppresses false alarms---pairing
BLEU/GLEU with BERTScore keeps semantics-preserving paraphrases from tripping the
\textsc{Description Changes} gate. Third, re-running our perturbation protocol on
a team's own goldens recalibrates the thresholds to its score distributions and
workflow sizes. Table~\ref{tab:bundle-recommendation} is thus a calibrated
starting point, not a fixed prescription.

\paragraph{Cost, latency, and open-weight judges.}
LLM-as-Judge dominates pipeline cost: at public GPT-4o pricing (\$2.50/M
input, \$10/M output tokens) each call runs $\sim$\$0.005--0.01 and seconds
of latency. At CI/CD scale this is non-trivial. The cheaper bundle metrics
can therefore pre-filter, reserving the judge for ambiguous
cases. \dataset{} also supports benchmarking open-weight judges and
learned workflow comparators.

\section{Conclusion}

We introduced \Dataset{}, a benchmark for evaluating workflow metrics under
controlled structural and textual perturbations, motivated by recurring
change-management failures in \textsc{LLexus}~\cite{lascasas2024llexus} over two
years of production operation. Our analysis shows distinct sensitivity patterns
and calibration gaps across failure modes, so no single uncalibrated score is a
sound basis for CI/CD shipping decisions.

We expect \Dataset{} to help teams operating workflow generators answer the same
question: whether a routine change has silently regressed a validated workflow.
Expressing goldens and candidates in the same node/edge form lets them apply the
calibrated bundle (Table~\ref{tab:bundle-recommendation}) as a CI/CD gate;
re-running our protocol on in-domain goldens recalibrates it.

\newpage
\section*{Limitations}

\textbf{Scope of evaluation.} \dataset{} evaluates workflows as static graph and
text artifacts: we measure how metric scores respond to controlled perturbations
but do not correlate calibration with downstream execution success, since no
shared, reproducible sandbox can faithfully execute workflows across the
heterogeneous domains and tool stacks in the corpus. We treat metric calibration
as a prerequisite for execution-aware evaluation, not a
substitute. We also study only three perturbation families
(\textsc{Missing Steps}, \textsc{Compressed Steps}, \textsc{Description Changes})
at three severities, chosen because each stresses a
different metric family. Step addition, duplication, reordering, wrong
dependencies, and wrong tool choice are not covered, so our calibration claims
may not extrapolate to them; each is a direct extension of the same protocol.

\textbf{Data and judge coverage.} The golden workflows are inherited from
WORFBENCH~\cite{qiao2024benchmarking} (English-only, filtered to $|V|\ge 5$) and
are concentrated at the short end (median 6 nodes; only 13 of 4{,}973 exceed 20).
Our size-stratified sensitivity analysis
(Appendix Table~\ref{tab:size-stratified}) is correspondingly thin at the top
end: the $16{+}$-node stratum contains just $22$ workflows, too few to establish
whether the calibration profile transfers to the longer, branch-rich workflows
that motivate the industrial framing; the $50\%$ setting also acts on very
small graphs for most of the corpus. Our LLM-as-Judge results use a
single judge (GPT-4o via Azure OpenAI at temperature~0); we do not benchmark
open-weight or alternative proprietary judges, and reproducibility depends on its
continued availability and version stability.

\textbf{Pipeline and metric choices.} Perturbed variants are produced by LLMs and
accepted by an automated validation pipeline with only spot-check human review,
so residual artifacts may persist in unreviewed portions of the 44{,}757-variant
corpus. Structural metrics (Graph/Chain F1) use a fixed alignment threshold
($\tau_{\text{align}}=0.8$) and encoder that were not swept; absolute scores would
shift under other settings, though the relative ordering used for sensitivity
(\S\ref{sec:sensitivity}) is the robust signal of interest. Finally, we scope the
study to seven widely deployed off-the-shelf metrics and do not compare against
learned or task-specific comparators, leaving that to future work.

\section*{Ethical Considerations}

\paragraph{Data provenance and licensing.}
The golden workflows underlying \dataset{} come from the training split of
WORFBENCH~\cite{qiao2024benchmarking} (Apache-2.0), whose English-language
workflows we use unchanged modulo the $|V|\ge 5$ filter; we redistribute
\dataset{} under the same license. All 44{,}757 perturbed variants are produced
synthetically by LLMs (GPT-4o, with GPT-4.1 and o3-mini fallbacks)
following the controlled protocol described in \S3. Four \textsc{Missing
Steps} rows were hand-corrected by the authors after automated regeneration
kept failing the static checks, and are flagged as manually authored. The
authors also checked $90$ variants ($10$ per perturbation type and severity)
for perturbation compliance; $89$ passed, with one $50\%$ Compressed-Steps
failure. That sample is drawn from WikiHow goldens and does not span all seven
sources; row-level decisions are in \texttt{manual\_inspection\_log.csv}. No
new human-subjects data was collected, and no personally identifying
information is present in the corpus.

\paragraph{Intended use and dual-use risks.}
\dataset{} is an offline diagnostic resource for calibrating evaluation
metrics used to validate LLM-generated workflows, for example inside CI/CD
regression tests. It does not
generate, execute, or recommend production actions; it scores existing
artifacts against a reference, so direct dual-use risk is low. A
practitioner relying \emph{solely} on an uncalibrated score for a production
ship/no-ship decision could ship a regression silently; the paper's central
finding (Section~\ref{sec:quant_results}) is precisely that this is
unsafe, hence the calibrated bundle in
Table~\ref{tab:bundle-recommendation} rather than any single score.

\paragraph{Compute, cost, and reproducibility.}
Reproducing the judge-based numbers depends on continued availability of
GPT-4o (used via Azure OpenAI at temperature~0) and is subject to provider
retirement policies, as noted in Limitations. All other metrics use open
implementations and are deterministic given the same input. Per-evaluation
cost estimates from public OpenAI pricing appear in \S6.

\paragraph{AI writing assistance.}
We used generative AI tools for polishing authored text. All technical
content, experimental design, analysis, and conclusions are the authors' own;
no AI tool was used to generate experimental results or citations.

\bibliography{references}

\begin{thebibliography}{22}
\providecommand{\natexlab}[1]{#1}

\bibitem[{Barke et~al.(2026)Barke, Goyal, Khare, Singh, Nath, and
  Bansal}]{barke2026agentrxdiagnosingaiagent}
Shraddha Barke, Arnav Goyal, Alind Khare, Avaljot Singh, Suman Nath, and Chetan
  Bansal. 2026.
\newblock \href {https://arxiv.org/abs/2602.02475} {{AgentRx}: Diagnosing {AI}
  agent failures from execution trajectories}.
\newblock \emph{Preprint}, arXiv:2602.02475.

\bibitem[{Bran et~al.(2024)Bran, Cox, Schilter, Baldassari, White, and
  Schwaller}]{bran2023chemcrowaugmentinglargelanguagemodels}
Andres~M Bran, Sam Cox, Oliver Schilter, Carlo Baldassari, Andrew~D White, and
  Philippe Schwaller. 2024.
\newblock \href {https://doi.org/10.1038/s42256-024-00832-8} {Augmenting large
  language models with chemistry tools}.
\newblock \emph{Nature Machine Intelligence}, 6(5):525--535.

\bibitem[{Chandrashekar(2025)}]{aws_amazonq_cli}
Kirankumar Chandrashekar. 2025.
\newblock \href
  {https://aws.amazon.com/blogs/devops/streamline-operational-troubleshooting-and-tasks-with-amazon-q-developer-cli/}
  {Streamline operational troubleshooting and tasks with {A}mazon {Q}
  {D}eveloper {CLI}}.
\newblock AWS DevOps Blog.
\newblock Posted 2025-06-19.

\bibitem[{Chiang and Lee(2023)}]{chiang2023largelanguagemodelsalternative}
Cheng-Han Chiang and Hung-yi Lee. 2023.
\newblock \href {https://doi.org/10.18653/v1/2023.acl-long.870} {Can large
  language models be an alternative to human evaluations?}
\newblock In \emph{Proceedings of the 61st Annual Meeting of the Association
  for Computational Linguistics (Volume 1: Long Papers)}, pages 15607--15631,
  Toronto, Canada. Association for Computational Linguistics.

\bibitem[{Dalal et~al.(2026)Dalal, Kanda, Wang, Ji, and
  Jain}]{dalal-etal-2026-compositional}
Dwip Dalal, Madhav Kanda, Zhenhailong Wang, Heng Ji, and Unnat Jain. 2026.
\newblock \href {https://doi.org/10.18653/v1/2026.findings-eacl.304}
  {Compositional reasoning via joint image and language decomposition}.
\newblock In \emph{Findings of the {A}ssociation for {C}omputational
  {L}inguistics: {EACL} 2026}, pages 5753--5775, Rabat, Morocco. Association
  for Computational Linguistics.

\bibitem[{Deshpande et~al.(2025)Deshpande, Gangal, Mehta, Krishnan, Kannappan,
  and Qian}]{deshpande2025trailtracereasoningagentic}
Darshan Deshpande, Varun Gangal, Hersh Mehta, Jitin Krishnan, Anand Kannappan,
  and Rebecca Qian. 2025.
\newblock \href {https://arxiv.org/abs/2505.08638} {{TRAIL}: Trace reasoning
  and agentic issue localization}.
\newblock \emph{Preprint}, arXiv:2505.08638.

\bibitem[{Jin et~al.(2024)Jin, Yang, Chen, and Lu}]{jin2024genegpt}
Qiao Jin, Yifan Yang, Qingyu Chen, and Zhiyong Lu. 2024.
\newblock \href {https://doi.org/10.1093/bioinformatics/btae075} {{GeneGPT}:
  augmenting large language models with domain tools for improved access to
  biomedical information}.
\newblock \emph{Bioinformatics}, 40(2):btae075.

\bibitem[{Kendall(1938)}]{Kendall1938}
Maurice~G. Kendall. 1938.
\newblock \href {https://doi.org/10.1093/biomet/30.1-2.81} {A new measure of
  rank correlation}.
\newblock \emph{Biometrika}, 30(1-2):81--93.

\bibitem[{Kuhn(1955)}]{kuhn1955hungarian}
Harold~W. Kuhn. 1955.
\newblock \href {https://doi.org/10.1002/nav.3800020109} {The {H}ungarian
  method for the assignment problem}.
\newblock \emph{Naval Research Logistics Quarterly}, 2(1--2):83--97.

\bibitem[{Las-Casas et~al.(2024)Las-Casas, Kumbhare, Fonseca, and
  Agarwal}]{lascasas2024llexus}
Pedro Las-Casas, Alok~Gautum Kumbhare, Rodrigo Fonseca, and Sharad Agarwal.
  2024.
\newblock \href {https://doi.org/10.1145/3689051.3689056} {{LLexus}: an {AI}
  agent system for incident management}.
\newblock \emph{ACM SIGOPS Operating Systems Review}, 58(1):47--60.

\bibitem[{{Microsoft}(2026{\natexlab{a}})}]{azure_openai_model_retirements}
{Microsoft}. 2026{\natexlab{a}}.
\newblock \href
  {https://learn.microsoft.com/en-us/azure/ai-foundry/openai/concepts/model-retirements}
  {{A}zure {OpenAI} in {A}zure {AI} {F}oundry {M}odels model retirements}.
\newblock Microsoft Learn documentation.

\bibitem[{{Microsoft}(2026{\natexlab{b}})}]{microsoft_powerautomate_copilot}
{Microsoft}. 2026{\natexlab{b}}.
\newblock \href
  {https://learn.microsoft.com/en-us/power-automate/create-cloud-flow-copilot}
  {Create a cloud flow in {P}ower {A}utomate using {C}opilot}.
\newblock Microsoft Learn documentation.
\newblock Last updated 2026-01-16.

\bibitem[{Pagonas et~al.(2026)Pagonas, Lou, Peng, Rubenstein, and
  Kaffes}]{pagonas2026vinelm}
Nikos Pagonas, Matthew Lou, Tianyi Peng, Dan Rubenstein, and Kostis Kaffes.
  2026.
\newblock \href {https://doi.org/10.48550/arXiv.2605.23914} {{VineLM}:
  Trie-based fine-grained control for agentic workflows}.
\newblock \emph{Computing Research Repository}.
\newblock ArXiv:2605.23914.

\bibitem[{Papineni et~al.(2002)Papineni, Roukos, Ward, and
  Zhu}]{papineni2002bleu}
Kishore Papineni, Salim Roukos, Todd Ward, and Wei-Jing Zhu. 2002.
\newblock \href {https://doi.org/10.3115/1073083.1073135} {{BLEU}: a method for
  automatic evaluation of machine translation}.
\newblock In \emph{Proceedings of the 40th Annual Meeting of the Association
  for Computational Linguistics}, pages 311--318, Philadelphia, Pennsylvania,
  USA. Association for Computational Linguistics.

\bibitem[{Patil et~al.(2025)Patil, Mao, Yan, Ji, Suresh, Stoica, and
  Gonzalez}]{patilberkeley}
Shishir~G Patil, Huanzhi Mao, Fanjia Yan, Charlie Cheng-Jie Ji, Vishnu Suresh,
  Ion Stoica, and Joseph~E Gonzalez. 2025.
\newblock \href {https://openreview.net/forum?id=2GmDdhBdDk} {The {B}erkeley
  {F}unction {C}alling {L}eaderboard ({BFCL}): From tool use to agentic
  evaluation of large language models}.
\newblock In \emph{Forty-second International Conference on Machine Learning}.

\bibitem[{Qiao et~al.(2025)Qiao, Fang, Qiu, Wang, Zhang, Jiang, Xie, Huang, and
  Chen}]{qiao2024benchmarking}
Shuofei Qiao, Runnan Fang, Zhisong Qiu, Xiaobin Wang, Ningyu Zhang, Yong Jiang,
  Pengjun Xie, Fei Huang, and Huajun Chen. 2025.
\newblock \href {https://openreview.net/forum?id=vunPXOFmoi} {Benchmarking
  agentic workflow generation}.
\newblock In \emph{The Thirteenth International Conference on Learning
  Representations}.

\bibitem[{Reimers and Gurevych(2019)}]{reimers-2019-sentence-bert}
Nils Reimers and Iryna Gurevych. 2019.
\newblock \href {https://doi.org/10.18653/v1/D19-1410} {Sentence-{BERT}:
  Sentence embeddings using {S}iamese {BERT}-networks}.
\newblock In \emph{Proceedings of the 2019 Conference on Empirical Methods in
  Natural Language Processing and the 9th International Joint Conference on
  Natural Language Processing (EMNLP-IJCNLP)}, pages 3982--3992, Hong Kong,
  China. Association for Computational Linguistics.

\bibitem[{Shi et~al.(2025)Shi, Ma, Liang, Diao, Ma, and
  Vosoughi}]{shi-etal-2025-judging}
Lin Shi, Chiyu Ma, Wenhua Liang, Xingjian Diao, Weicheng Ma, and Soroush
  Vosoughi. 2025.
\newblock \href {https://doi.org/10.18653/v1/2025.ijcnlp-long.18} {Judging the
  judges: A systematic study of position bias in {LLM}-as-a-judge}.
\newblock In \emph{Proceedings of the 14th International Joint Conference on
  Natural Language Processing and the 4th Conference of the Asia-Pacific
  Chapter of the Association for Computational Linguistics}, pages 292--314,
  Mumbai, India. The Asian Federation of Natural Language Processing and The
  Association for Computational Linguistics.

\bibitem[{Wang et~al.(2024)Wang, Li, Chen, Cai, Zhu, Lin, Cao, Kong, Liu, Liu,
  and Sui}]{wang2023largelanguagemodelsfair}
Peiyi Wang, Lei Li, Liang Chen, Zefan Cai, Dawei Zhu, Binghuai Lin, Yunbo Cao,
  Lingpeng Kong, Qi~Liu, Tianyu Liu, and Zhifang Sui. 2024.
\newblock \href {https://doi.org/10.18653/v1/2024.acl-long.511} {Large language
  models are not fair evaluators}.
\newblock In \emph{Proceedings of the 62nd Annual Meeting of the Association
  for Computational Linguistics (Volume 1: Long Papers)}, pages 9440--9450,
  Bangkok, Thailand. Association for Computational Linguistics.

\bibitem[{Wu et~al.(2016)Wu, Schuster, Chen, Le, Norouzi, Macherey, Krikun,
  Cao, Gao, Macherey et~al.}]{wu2016google}
Yonghui Wu, Mike Schuster, Zhifeng Chen, Quoc~V. Le, Mohammad Norouzi, Wolfgang
  Macherey, Maxim Krikun, Yuan Cao, Qin Gao, Klaus Macherey, and 1 others.
  2016.
\newblock \href {https://arxiv.org/abs/1609.08144} {Google's neural machine
  translation system: Bridging the gap between human and machine translation}.
\newblock \emph{Preprint}, arXiv:1609.08144.

\bibitem[{Zhang et~al.(2020)Zhang, Kishore, Wu, Weinberger, and
  Artzi}]{zhang2020bertscoreevaluatingtextgeneration}
Tianyi Zhang, Varsha Kishore, Felix Wu, Kilian~Q. Weinberger, and Yoav Artzi.
  2020.
\newblock \href {https://arxiv.org/abs/1904.09675} {{BERTS}core: Evaluating
  text generation with {BERT}}.
\newblock In \emph{International Conference on Learning Representations}.

\bibitem[{Zheng et~al.(2023)Zheng, Chiang, Sheng, Zhuang, Wu, Zhuang, Lin, Li,
  Li, Xing, Zhang, Gonzalez, and
  Stoica}]{zheng2023judgingllmasajudgemtbenchchatbot}
Lianmin Zheng, Wei-Lin Chiang, Ying Sheng, Siyuan Zhuang, Zhanghao Wu, Yonghao
  Zhuang, Zi~Lin, Zhuohan Li, Dacheng Li, Eric~P. Xing, Hao Zhang, Joseph~E.
  Gonzalez, and Ion Stoica. 2023.
\newblock \href
  {https://proceedings.neurips.cc/paper_files/paper/2023/hash/91f18a1287b398d378ef22505bf41832-Abstract-Datasets_and_Benchmarks.html}
  {Judging {LLM}-as-a-judge with {MT}-{B}ench and {C}hatbot {A}rena}.
\newblock In \emph{Advances in Neural Information Processing Systems 36
  (NeurIPS 2023) Track on Datasets and Benchmarks}.

\end{thebibliography}
\appendix
\label{sec:appendix}
\newpage

\section*{Appendix}

In this appendix, we include the following to supplement the main paper:

\begin{enumerate}[ label=\textbf{\Alph*.}, leftmargin=*, align=left ]
    \item \textbf{Prompt Templates and Generation Pipeline}
    (Section~\ref{appendix:prompts}). Exact LLM
    prompts used to generate perturbed workflow variants, including
    error-injection and refinement strategies, the static-check failure
    trajectory and residual-failure handling, and the description-change
    generation pipeline.

    \item \textbf{Detailed Metric Definitions} (Section~\ref{appendix:metrics}).
    Formal definitions, equations, and implementation details for all evaluation
    metrics.

    \item \textbf{Extended Results and Sensitivity Analysis}
    (Section~\ref{appendix:results}). Complete metric tables across perturbation
    types and a detailed analysis of metric sensitivity under controlled
    degradation.
\end{enumerate}

\section{Prompt Templates and Generation Pipeline}
\label{appendix:prompts}

The following prompts were used to generate perturbed workflow variants.
Placeholders $N$ denote the exact number of nodes to alter, computed before the
prompt is assembled as $N=\max(1,\lceil |V|\times\alpha\rceil)$ for severity
$\alpha\in\{0.1,0.3,0.5\}$ (see \S\ref{sec:benchmark-construction}). The prompts
below show the instruction template; each call also embeds the concrete golden
workflow.

\subsection{Few-Shot Examples (prepended to every generator call)}
\label{appendix:few-shot}

For each perturbation type, every generator call prepends a fixed few-shot
example before the row-specific instruction. The row-specific portion consists of
the original golden workflow plus the perturbation directive, in the form shown in
Section~\ref{appendix:row-prompts} below.

\paragraph{Missing Steps.}
\begin{quote}\small\ttfamily
Original workflow:\\
Node:\\
1: go to potential locations where a kettle may be found\\
2: take kettle from where it was found\\
\ldots\\
7: put kettle in/on storage.\\
Edge: (START,1) (1,2) \ldots (6,7) (7,END)\\
Please remove one step from the workflow above and return the new workflow in the same Node:/Edge: format.\\
Perturbed (one step missing):\\
Node:\\
1: go to potential locations where a kettle may be found\\
\ldots\\
6: open storage if necessary\\
Edge: (START,1) (1,2) \ldots (5,6) (6,END)\\
---
\end{quote}

\paragraph{Compressed Steps.}
\begin{quote}\small\ttfamily
Original workflow:\\
Node:\\
1: locate the mug in the kitchen\\
2: pick up the mug\\
3: go to the coffee machine\\
4: place the mug under the coffee spout\\
5: press the button to brew coffee\\
Edge: (START,1) (1,2) (2,3) (3,4) (4,5) (5,END)\\
Please compress two steps in the workflow above (combine them into fewer steps) and return the new workflow in the same Node:/Edge: format.\\
Perturbed (two steps compressed):\\
Node:\\
1: locate the mug in the kitchen and pick it up\\
2: go to the coffee machine\\
3: place the mug under the coffee spout and press the button to brew coffee\\
Edge: (START,1) (1,2) (2,3) (3,END)\\
---
\end{quote}

\paragraph{Description Changes.}
\begin{quote}\small\ttfamily
Original workflow:\\
Node:\\
1: go to the living room\\
2: find the remote control\\
3: turn on the television\\
4: select the news channel\\
Edge: (START,1) (1,2) (2,3) (3,4) (4,END)\\
Please paraphrase exactly two steps in the workflow above (change their description but not their meaning). Do not change more or fewer than two steps. Keep all other node descriptions exactly the same. Return the new workflow in the same Node:/Edge: format.\\
Perturbed (two steps paraphrased):\\
Node:\\
1: go to the living room\\
2: locate the remote control device\\
3: turn on the television\\
4: switch to the news channel\\
Edge: (START,1) (1,2) (2,3) (3,4) (4,END)\\
---
\end{quote}

\subsection{Row-Specific Prompt Templates}
\label{appendix:row-prompts}

In the templates below, \{golden workflow\} denotes the concrete golden workflow
inserted into the prompt, and $N$ is the exact number of nodes to alter, defined
above.

\subsection{Missing Steps Prompt}
\begin{quote}
\texttt{Original workflow: \{golden workflow\}}\\
\texttt{Please remove exactly $N$ steps from the workflow above and return the new workflow in the same Node:/Edge: format.}\\
\texttt{Perturbed ($N$ steps missing):}
\end{quote}

\subsection{Compressed Step Prompt}
\begin{quote}
\texttt{Original workflow: \{golden workflow\}}\\
\texttt{Please compress exactly $N$ steps in the workflow above (combine them into fewer steps) and return the new workflow in the same Node:/Edge: format.}\\
\texttt{Perturbed ($N$ steps compressed):}
\end{quote}

\subsection{Description Changes Prompt}
\begin{quote}
\texttt{Original workflow: \{golden workflow\}}\\
\texttt{Please paraphrase exactly $N$ steps in the workflow above (change their description but not their meaning). Do not change more or fewer than $N$ steps. Keep all other node descriptions exactly the same. Return the new workflow in the same Node:/Edge: format.}\\
\texttt{Perturbed ($N$ steps paraphrased):}
\end{quote}

\subsection{Iterative Refinement and Escalation}
When a generated output fails validation, the variant enters a feedback-driven
refinement loop in which the validation error is re-injected into the next
prompt. For example:
\begin{quote}
\texttt{Your output had 8 nodes instead of 9 for a 10\% removal. Please
regenerate the workflow and ensure that exactly one node is removed. Return only
the modified workflow as a numbered list of steps.}
\end{quote}
If the corrected output again fails validation (e.g., paraphrases 4 nodes
instead of 3), the process is repeated up to three times per model. After three
unsuccessful iterations, the task is escalated to a stronger generation model.

\subsection{Residual-Failure Handling}
\label{appendix:pipeline}

The body reports the static-check failure trajectory of the iterative refinement
loop. The raw initial-pass counts are $2{,}634 / 14{,}919$ for Missing Steps,
$4{,}106 / 14{,}919$ for Compressed Steps, and $6{,}673 / 14{,}919$ for
Description Changes, and the successive fallbacks were GPT-4.1, an Azure-OpenAI
content filter, and finally o3-mini with feedback-driven re-prompting. The few
residual variants that survived were then resolved per split. These are
generation-time fallback counts, and are distinct from the final-release
node-count audit reported below. The $4$ residual
\textsc{Missing Steps} variants that continued to fail automated regeneration
were hand-corrected by the authors and are flagged as manually authored in the
released dataset. The $21$ residual \textsc{Compressed Steps} variants were
blocked by an Azure OpenAI content filter on the perturbed workflow text and were
regenerated through a standard OpenAI API endpoint. The $200$ residual
\textsc{Description Changes} variants were repaired by the hybrid fallback in
Appendix~\ref{appendix:descpipeline}.

\subsection{Final-Release Node-Count Audit}
\label{appendix:nodeaudit}

A recount of the released dataset flagged $175$ node-count mismatches: $100$
\textsc{Missing Steps}, $50$ \textsc{Compressed Steps}, and $25$
\textsc{Description Changes}. Among \textsc{Missing Steps}, $98$ variants
contain one extra node, one contains one fewer, and one contains two extra; all
$50$ \textsc{Compressed Steps} mismatches contain one extra node. Of the $25$
\textsc{Description Changes} flags, $18$ arise from six source goldens whose
stored node-count metadata disagrees with their parsed node lists, and seven
from variant-level node-list or formatting defects. All flagged rows are
retained and marked in the release rather than dropped; excluding them shifts
no per-severity metric mean by more than $0.0022$.

\subsection{Manual Compliance Inspection}
\label{appendix:inspection}

Separately from the automated validation, we manually inspected $10$ variants
at each severity for each perturbation type ($90$ total) against type-specific
criteria: deletion removed the intended number of steps without adding content,
compression retained the original actions, and paraphrasing preserved meaning
and topology. This was a perturbation-compliance check, not a human
severity-rating study. The sole failure was
the UPS entry identified by \texttt{perturbation\_type=compressed\_steps},
\texttt{perturbation\_percent=50}, and
\texttt{manual\_inspection\_sample=7} in
\texttt{manual\_inspection\_log.csv}.
The compressed workflow retained the instruction to print the shipping label
but omitted the subsequent action to send the package or drop it off at a UPS
location. Because that action was lost rather than merged into another step, we
marked the entry as a failure. The inspected subset therefore achieved an
overall perturbation-compliance rate of $89/90$ ($98.9\%$). The $90$ inspected
variants are drawn from WikiHow-sourced goldens and do not span all seven
source datasets; broadening the inspection across sources is left to future
work.

\subsection{Description-Change Generation Pipeline}
\label{appendix:descpipeline}

Description-change variants follow the same primary path as the other
perturbation types: a bulk workflow-editing prompt (few-shot plus row-specific
instruction) asks the LLM to paraphrase exactly $N$ step descriptions while
preserving graph structure, with up to three static-check retries and subsequent
escalation to stronger models with feedback-driven re-prompting. For the
$\approx 200$ variants that still failed after escalation, we applied a
\emph{hybrid} fallback: $N$ node indices are selected at random, a generator LLM
paraphrases each selected step description in a separate call, and the workflow
is reassembled with edges unchanged. This fallback is more reliable than
single-shot whole-workflow editing.

\section{Detailed Metric Definitions}
\label{appendix:metrics}

We present the formal definitions and implementation details of all evaluation
metrics used in \Dataset. Where applicable, we include equations and
illustrative examples showing how the scores are computed.

\subsection{Chain F1}
\label{appendix:chain-f1}
\paragraph{Idealized objective.}
Suppose the predicted node chain is \(C(V^{p})\) and the gold workflow graph is
\(G(V^{g}, E^{g})\). From \(G\), we can enumerate all possible topological
orderings \(\{C(V^{g})_1, C(V^{g})_2, \ldots \}\). The predicted chain is
compared against each ordering using the \emph{Longest Increasing Subsequence
(LIS)}:

\[
l_i = \text{LIS}(C(V^{g})_i, C(V^{p})),
\]

and the longest valid subsequence length is taken as
\[
l = \max(|l_1|, |l_2|, \ldots, |l_n|).
\]

Precision and recall are then
\[
p_{\text{chain}} = \frac{l}{|V^{p}|}, \qquad r_{\text{chain}} = \frac{l}{|V^{g}|},
\]

yielding
\[
f_{1}^{\text{chain}} = \frac{2 \, p_{\text{chain}} \, r_{\text{chain}}}{p_{\text{chain}} + r_{\text{chain}}}.
\]

\paragraph{Implementation.}
Enumerating every topological ordering of \(G\) is exponential in
$|V^{g}|$ in the worst case, and is further complicated by the fact that node
identities are not shared across workflows. The predicted and gold node sets
differ in surface form and must first be aligned semantically. We therefore use
the following polynomial-time approximation:

\begin{enumerate}[leftmargin=*, itemsep=2pt, parsep=0pt, topsep=2pt]
  \item Embed each step description (gold and predicted) with
  Sentence-BERT~\cite{reimers-2019-sentence-bert}, using the
  \texttt{all-mpnet-base-v2} encoder.
  \item Compute the cosine-similarity matrix between predicted and gold step
  embeddings, and solve an optimal one-to-one assignment with the Hungarian
  algorithm~\cite{kuhn1955hungarian}
  (\texttt{scipy.optimize.linear\_sum\_assignment}).
  \item Retain only assigned pairs whose cosine similarity is
  $\geq \tau_{\text{align}} = 0.8$; unmatched predicted steps are dropped.
  \item Let $\sigma$ be the sequence of gold indices that the retained
  predicted steps map to, taken in predicted-chain order. Compute
  $l = |\text{LIS}(\sigma)|$ under the natural gold ordering.
\end{enumerate}

This realizes the LIS objective against a \emph{single} canonical gold ordering
(the order in which the gold graph's steps are listed), rather than against
all topological orderings. The substitution is exact for the linear-chain
workflows that dominate our corpus (where the topological ordering is unique)
and is an approximation for the small number of multi-branch workflows.

\paragraph{Threshold and encoder.}
Both the encoder choice and the alignment threshold $\tau_{\text{align}}$ are
fixed across all experiments to keep relative comparisons across perturbations
meaningful. The threshold of $0.8$ was chosen because in the
\texttt{all-mpnet-base-v2} embedding space sentence pairs with cosine
similarity below this value are typically only loosely related, while values
above it correspond to paraphrastic or near-paraphrastic agreement. We did not
run a full sweep over $\tau_{\text{align}}$; this dependence is acknowledged in
the Limitations section.

\subsection{Graph F1}
\label{appendix:graph-f1}
\paragraph{Idealized objective.}
Given predicted graph \(G(V^{p}, E^{p})\) and gold graph \(G(V^{g}, E^{g})\), we
seek the Maximum Common Induced Subgraph (MCIS):

\[
\begin{aligned}
G_{mcis}(V_{mcis}, E_{mcis})
&= \text{MCIS}\!\Big(
    G(V^{p}, E^{p}), \\
&\qquad\qquad\;\; G(V^{g}, E^{g})
\Big).
\end{aligned}
\]

If the largest matched subgraph contains \(k = |V_{mcis}|\) nodes, then

\[
p_{\text{graph}} = \frac{k}{|V^{p}|}, \qquad r_{\text{graph}} = \frac{k}{|V^{g}|},
\]

and the final score is
\[
f_{1}^{\text{graph}} = \frac{2 \, p_{\text{graph}} \, r_{\text{graph}}}{p_{\text{graph}} + r_{\text{graph}}}.
\]

This procedure captures both node correctness and edge consistency, providing a
stricter evaluation than simple edge overlap.

\paragraph{Implementation.}
Exact MCIS on labeled graphs is NP-hard. As in Chain F1, the
predicted and gold node sets do not share identifiers and must be aligned
semantically before any subgraph notion makes sense. We approximate the MCIS
objective as follows:

\begin{enumerate}[leftmargin=*, itemsep=2pt, parsep=0pt, topsep=2pt]
  \item Compute the same SBERT (\texttt{all-mpnet-base-v2}) embeddings,
  cosine-similarity matrix, and Hungarian one-to-one assignment as in
  Chain F1, and retain only pairs with cosine similarity
  $\geq \tau_{\text{align}} = 0.8$. Call the retained alignment $\pi$ and let
  $S = \pi(V^{p}) \subseteq V^{g}$.
  \item Map every predicted edge $(u,v) \in E^{p}$ through $\pi$, producing a
  set of induced edges on $S$.
  \item Iteratively prune $S$: remove any node $v \in S$ whose neighbor set
  induced from the mapped predicted edges within $S$ does not equal its
  neighbor set in the gold graph restricted to $S$. Repeat until no further
  removal is possible.
  \item Let $k = |S|$ after pruning and compute $p_{\text{graph}}$,
  $r_{\text{graph}}$, $f_{1}^{\text{graph}}$ as above.
\end{enumerate}

The pruning step enforces the \emph{induced-subgraph} condition: a node is
retained only if its neighborhood within $S$ is identical under both the gold
and the mapped predicted edge sets. The output is therefore a common induced
subgraph of $G^{p}$ and $G^{g}$ under the alignment $\pi$. It is not necessarily
the maximum one, since $\pi$ is fixed up front by the embedding-based
assignment rather than searched jointly with $S$. The same encoder and
threshold as in Chain F1 apply, with the same caveat about $\tau_{\text{align}}$
sensitivity.

\subsection{BLEU and GLEU}
BLEU~\cite{papineni2002bleu} measures $n$-gram precision between candidate and
gold step descriptions, with a brevity penalty to discourage very short outputs.
GLEU~\cite{wu2016google} balances precision and recall over $n$-grams, making it
more robust to short sequences. In our implementation, all step descriptions of
a workflow are concatenated (whitespace-joined, in order) into a single
document, and the score is computed once per workflow against the corresponding
gold concatenated document using NLTK's \texttt{sentence\_bleu} (with
\texttt{SmoothingFunction().method1}) and \texttt{sentence\_gleu}, with
\texttt{nltk.word\_tokenize} tokenization. Workflow-level scores are then
averaged across the dataset to produce the means reported in
Table~\ref{tab:sensitivity-summary} and Appendix~\ref{appendix:results}.

\subsection{BERTScore}
BERTScore~\cite{zhang2020bertscoreevaluatingtextgeneration} measures the
semantic similarity of step descriptions using contextual token embeddings. Our
implementation calls the public \texttt{bert\_score}
library\footnote{\url{https://github.com/Tiiiger/bert_score}} with
\texttt{lang="en"} and \texttt{rescale\_with\_baseline=True}; under these
settings the library selects \texttt{roberta-large} as the encoder and rescales
each token-level similarity by the corpus-baseline similarity between
unrelated sentence pairs. We report the rescaled F1, averaged across the
aligned step pairs within a workflow and then across the dataset. The baseline
rescaling makes the absolute scores noticeably lower than the un-rescaled
BERTScore values typically reported in the MT/summarization literature; the
quantity of interest in this paper is the \emph{relative} change in BERTScore
across perturbation severities, which is unaffected by the shift.

\paragraph{Alignment of unequal step lists.}
Since perturbations such as \emph{Missing Steps} and \emph{Compressed Steps}
change the number of steps, the gold and perturbed step lists are not the same
length. BERTScore is fundamentally a paired-list metric (item~$i$ on the
candidate side is compared with item~$i$ on the reference side), so some
alignment policy is required before it can be invoked. We use a deliberately
simple and conservative rule: order-preserving truncation to
$\min(|gold|, |pred|)$, comparing the first $\min$ step descriptions on each
side. This choice keeps BERTScore as a \emph{pure semantic-similarity}
signal that complements, rather than duplicates, the other metrics in our
bundle. Specifically: (i) the cost of \emph{deletions} is already accounted for
by the structural metrics (Graph and Chain F1), so introducing a
length-mismatch penalty inside BERTScore would double-count missing steps; and
(ii) optimal-assignment alignments (e.g., Hungarian matching over embedding
similarity) would re-import the very signal that node-cosine similarity and the
embedding-based Kendall's $\tau$ alignment in the bundle are designed to
provide, again collapsing the complementarity between metrics. Since our
perturbations operate on contiguous subsets and preserve the order of the
remaining steps, the truncated prefix is a faithful slice of the perturbed
workflow for semantic comparison.

\subsection{Kendall's $\tau$}
Kendall's $\tau$~\cite{Kendall1938} measures the ordinal correlation between two
ranked lists. For workflows, we treat the step ordering as a ranking and compute
\[
\tau = \frac{C - D}{\tfrac{1}{2} n(n-1)},
\]
where $n$ is the number of aligned steps being compared, and $C$ and $D$ denote
the numbers of concordant and discordant step pairs, respectively. $\tau$ ranges
from $-1$ (complete disagreement) to $1$ (perfect agreement).

\subsection{LLM-as-Judge Prompts}
\label{appendix:llm-prompts}

For the LLM-as-Judge evaluation, we designed a rubric-based prompt that
instructs the model to compare a generated workflow against a gold reference.  
Workflows are formatted with explicit node and edge lists.  

\subsubsection{Rubric}
\begin{itemize}
  \item 0 = Completely Incorrect or Irrelevant  
  \item 1 = Major Flaws (breaks core logic)  
  \item 2 = Significant Issues (omits several key steps or adds wrong steps)  
  \item 3 = Noticeable Errors (minor reorder or one extra/missing non-critical
  step)  
  \item 4 = Very Good (only one minor step missing)  
  \item 5 = Perfect Match (fully adheres, detailed, accurate)  
\end{itemize}

\subsubsection{Anchor Examples}
We provide concrete anchors to calibrate the model's scoring:
\begin{itemize}
  \item Example A (Score 4): Workflow missing one minor step.  
  \item Example B (Score 5): Workflow perfectly matches the gold.  
  \item Example C (Score 3): Workflow with minor reordering of steps.  
  \item Example D (Score 2): Workflow with extra irrelevant steps.  
  \item Example E (Score 0): Workflow completely incorrect relative to task.  
\end{itemize}

\subsubsection{Prompt Format}
The model is instructed to:
\begin{enumerate}
  \item List nodes and edges from both workflows.  
  \item Match nodes by content.  
  \item Identify missing, extra, or reordered steps.  
  \item Assign an initial score using the rubric.  
  \item Perform a self-check: confirm if the initial score strictly follows the
  rubric.  
  \item Adjust to a final score if necessary.  
\end{enumerate}

\subsubsection{Expected Output}
The model must return exactly one JSON object with the fields:
{\footnotesize
\begin{verbatim}
{
  "ThoughtChain": "<step-by-step reasoning>",
  "initial_score": <int 0-5>,
  "self_check": {
    "consistent": "Yes"|"No",
    "final_score": <int 0-5>
  },
  "justification": "<one-sentence summary>"
}
\end{verbatim}
}

\subsubsection{Implementation Notes}
We use GPT-4o via Azure OpenAI with temperature~$0.0$ to minimize randomness.
Each workflow pair is evaluated by three independent calls of the same prompt
template;
the three per-call scores are averaged to absorb residual non-determinism from
the inference backend. We do not vary the prompt template across calls.

\section{Results Analysis}
\label{appendix:results}

\begin{figure*}[t]
  \centering
  \includegraphics[width=0.8\linewidth]{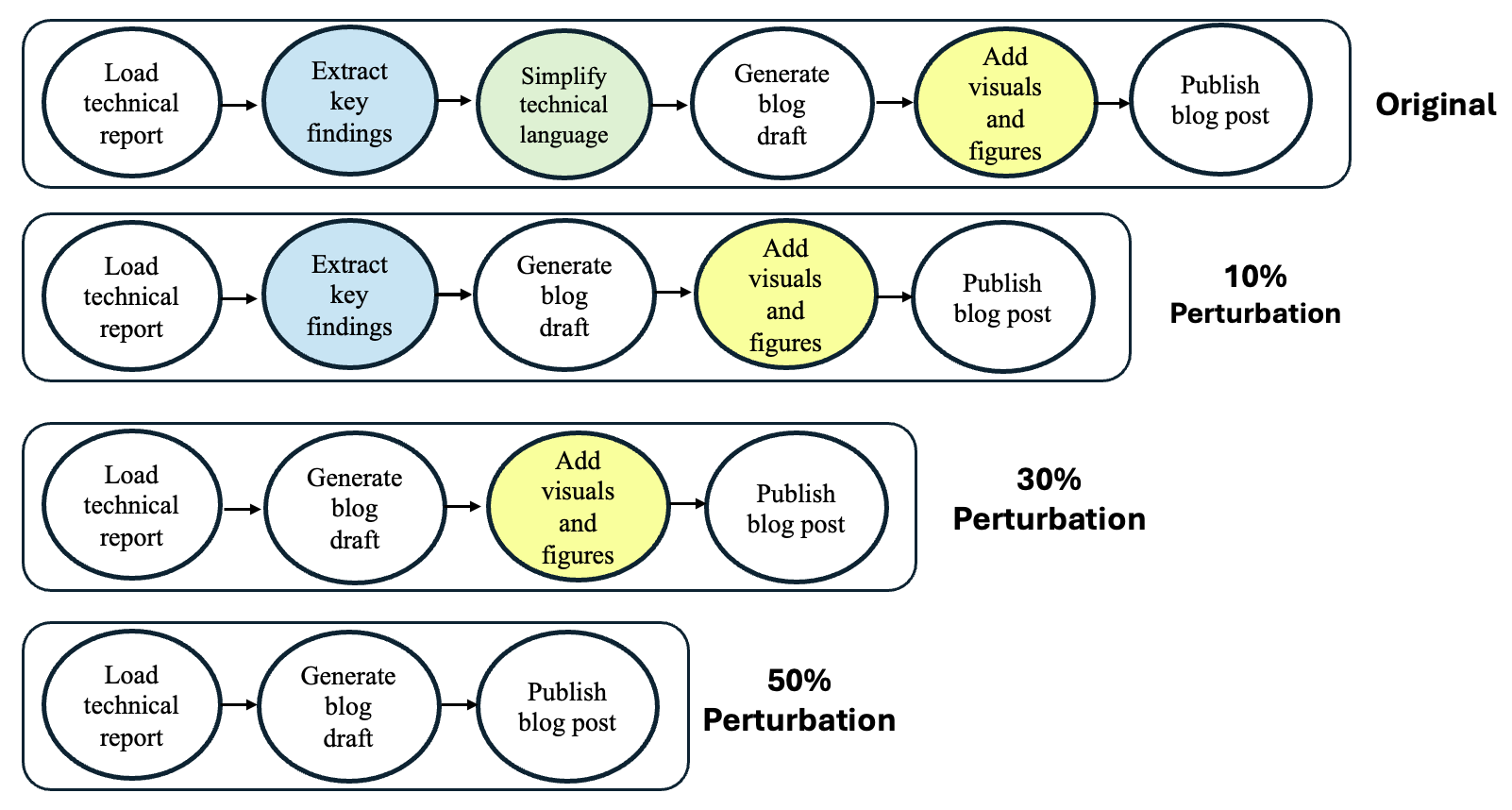}
  \caption{Visualization of \textsc{Missing Steps} perturbations applied to the
  six-node blog generation workflow. At each level, a proportional number of
  nodes is randomly dropped: 1 node (10\%), 2 nodes (30\%), and 3 nodes (50\%).
  Higher perturbation levels result in increasingly incomplete workflows that
  diverge from the golden sequence.}
  \label{fig:diff-percent-perturb}
\end{figure*}

\begin{table}[t]
  \centering
  \caption{Average metric scores (mean $\pm$ std) at different perturbation levels for Missing Steps.}
  \label{tab:score_change_missing}

  \setlength{\tabcolsep}{4pt}
  \renewcommand{\arraystretch}{1.08}
  \small

  \resizebox{\columnwidth}{!}{%
  \begin{tabular}{@{}lccc@{}}
    \toprule
    \textbf{Metric} & \multicolumn{3}{c}{\textbf{Perturbation Level}} \\
    \cmidrule(lr){2-4} & \textbf{10\%} & \textbf{30\%} & \textbf{50\%} \\
    \midrule
    Graph F1        & $0.90\!\pm\!0.06$ & $0.76\!\pm\!0.06$ & $0.61\!\pm\!0.06$
    \\
    Chain F1        & $0.90\!\pm\!0.07$ & $0.76\!\pm\!0.06$ & $0.61\!\pm\!0.06$
    \\
    BLEU            & $0.79\!\pm\!0.13$ & $0.52\!\pm\!0.14$ & $0.29\!\pm\!0.13$
    \\
    GLEU            & $0.80\!\pm\!0.11$ & $0.58\!\pm\!0.11$ & $0.42\!\pm\!0.11$
    \\
    BERTScore       & $0.81\!\pm\!0.26$ & $0.68\!\pm\!0.28$ & $0.72\!\pm\!0.30$
    \\
    Kendall's Tau   & $0.81\!\pm\!0.09$ & $0.61\!\pm\!0.08$ & $0.44\!\pm\!0.07$
    \\
    LLM-as-Judge & $0.64\!\pm\!0.21$ & $0.36\!\pm\!0.08$ & $0.32\!\pm\!0.10$ \\
    \bottomrule
  \end{tabular}%
  }
\end{table}

\begin{table}[t]
  \centering
  \caption{Average metric scores (mean $\pm$ std) at different perturbation levels for Compressed Steps.}
  \label{tab:score_change_compress}

  \setlength{\tabcolsep}{4pt}
  \renewcommand{\arraystretch}{1.08}
  \small

  \resizebox{\columnwidth}{!}{%
  \begin{tabular}{@{}lccc@{}}
    \toprule
    \textbf{Metric} & \multicolumn{3}{c}{\textbf{Perturbation Level}} \\
    \cmidrule(lr){2-4} & \textbf{10\%} & \textbf{30\%} & \textbf{50\%} \\
    \midrule
    Graph F1        & $0.86\!\pm\!0.10$ & $0.66\!\pm\!0.15$ & $0.45\!\pm\!0.18$
    \\
    Chain F1        & $0.86\!\pm\!0.10$ & $0.66\!\pm\!0.15$ & $0.44\!\pm\!0.18$
    \\
    BLEU            & $0.87\!\pm\!0.14$ & $0.71\!\pm\!0.17$ & $0.51\!\pm\!0.20$
    \\
    GLEU            & $0.86\!\pm\!0.12$ & $0.70\!\pm\!0.15$ & $0.53\!\pm\!0.17$
    \\
    BERTScore       & $0.41\!\pm\!0.26$ & $0.36\!\pm\!0.25$ & $0.32\!\pm\!0.24$
    \\
    Kendall's Tau   & $0.74\!\pm\!0.14$ & $0.46\!\pm\!0.16$ & $0.17\!\pm\!0.19$
    \\
   LLM-as-Judge & $0.65\!\pm\!0.22$ & $0.41\!\pm\!0.13$ & $0.37\!\pm\!0.10$ \\
    \bottomrule
  \end{tabular}%
  }
\end{table}

\begin{table}[t]
  \centering
  \caption{Average metric scores (mean $\pm$ std) at different perturbation levels for Description Changes.}
  \label{tab:score_change_desc}

  \setlength{\tabcolsep}{4pt}
  \renewcommand{\arraystretch}{1.08}
  \small

  \resizebox{\columnwidth}{!}{%
  \begin{tabular}{@{}lccc@{}}
    \toprule
    \textbf{Metric} & \multicolumn{3}{c}{\textbf{Perturbation Level}} \\
    \cmidrule(lr){2-4} & \textbf{10\%} & \textbf{30\%} & \textbf{50\%} \\
    \midrule
    Graph F1        & $0.97\!\pm\!0.08$ & $0.91\!\pm\!0.12$ & $0.85\!\pm\!0.16$
    \\
    Chain F1        & $0.96\!\pm\!0.08$ & $0.91\!\pm\!0.12$ & $0.85\!\pm\!0.16$
    \\
    BLEU            & $0.85\!\pm\!0.12$ & $0.66\!\pm\!0.14$ & $0.50\!\pm\!0.15$
    \\
    GLEU            & $0.86\!\pm\!0.10$ & $0.67\!\pm\!0.13$ & $0.52\!\pm\!0.14$
    \\
    BERTScore       & $0.94\!\pm\!0.20$ & $0.86\!\pm\!0.21$ & $0.79\!\pm\!0.21$
    \\
    Kendall's Tau   & $1.00\!\pm\!0.05$ & $0.99\!\pm\!0.08$ & $0.99\!\pm\!0.09$
    \\
    LLM-as-Judge & $0.98\!\pm\!0.07$ & $0.99\!\pm\!0.07$ & $0.99\!\pm\!0.05$ \\
    \bottomrule
  \end{tabular}%
  }
\end{table}

\begin{table}[t]
  \centering
  \small
  \setlength{\tabcolsep}{4pt}
  \renewcommand{\arraystretch}{1.05}
  \caption{Metric sensitivity $\Delta_m^{\mathrm{avg}}$ (per-row mean degradation
  rate, Section~\ref{sec:sensitivity}) across all metrics and perturbation types.
  Markers: $\blacktriangle$~high ($\ge 0.8$),
  $\bullet$~moderate ($0.3$--$0.8$),
  \,$\circ$\,~low ($<0.3$).}
  \label{tab:sensitivity-summary}
  \begin{tabular}{@{}lccc@{}}
    \toprule
    \textbf{Metric} & \textbf{Missing} & \textbf{Compressed} & \textbf{Description} \\
    \midrule
    Graph F1          & $0.72\,\bullet$       & $1.04\,\blacktriangle$ & $0.29\,\circ$ \\
    Chain F1          & $0.72\,\bullet$       & $1.04\,\blacktriangle$ & $0.29\,\circ$ \\
    BLEU              & $1.23\,\blacktriangle$ & $0.88\,\blacktriangle$ & $0.88\,\blacktriangle$ \\
    GLEU              & $0.96\,\blacktriangle$ & $0.84\,\blacktriangle$ & $0.84\,\blacktriangle$ \\
    BERTScore         & $0.24\,\circ$         & $0.23\,\circ$         & $0.39\,\bullet$ \\
    Kendall's $\tau$  & $0.94\,\blacktriangle$ & $1.42\,\blacktriangle$ & $0.02\,\circ$   \\
    LLM-as-Judge      & $0.80\,\blacktriangle$ & $0.70\,\bullet$       & $-0.03\,\circ$  \\
    \bottomrule
  \end{tabular}
\end{table}

\begin{table}[t]
  \centering
  \small
  \setlength{\tabcolsep}{3pt}
  \renewcommand{\arraystretch}{1.05}
  \caption{Two-slope decomposition of the sensitivity statistic: the
  mild$\to$moderate slope $\Delta_m^{1}=[\bar{s}_m(10\%)-\bar{s}_m(30\%)]/0.20$ and
  the moderate$\to$severe slope $\Delta_m^{2}=[\bar{s}_m(30\%)-\bar{s}_m(50\%)]/0.20$,
  whose average is the $\Delta_m^{\mathrm{avg}}$ of
  Table~\ref{tab:sensitivity-summary}. Divergent slopes---which
  $\Delta_m^{\mathrm{avg}}$ averages away---are shown in bold: BERTScore turns
  \emph{non-monotonic} under Missing Steps, and LLM-as-Judge is strongly
  front-loaded under both structural modes.}
  \label{tab:two-slope}
  \begin{tabular}{@{}l cc cc cc@{}}
    \toprule
    & \multicolumn{2}{c}{\textbf{Missing}}
    & \multicolumn{2}{c}{\textbf{Compressed}}
    & \multicolumn{2}{c}{\textbf{Description}} \\
    \cmidrule(lr){2-3}\cmidrule(lr){4-5}\cmidrule(lr){6-7}
    \textbf{Metric} & $\Delta^{1}$ & $\Delta^{2}$ & $\Delta^{1}$ & $\Delta^{2}$
                    & $\Delta^{1}$ & $\Delta^{2}$ \\
    \midrule
    Graph F1         & 0.71 & 0.74 & 1.00 & 1.08 & 0.27 & 0.30 \\
    Chain F1         & 0.71 & 0.74 & 1.01 & 1.07 & 0.28 & 0.30 \\
    BLEU             & 1.33 & 1.14 & 0.80 & 0.95 & 0.97 & 0.80 \\
    GLEU             & 1.08 & 0.84 & 0.79 & 0.88 & 0.93 & 0.76 \\
    BERTScore        & \textbf{0.67} & \textbf{$-0.20$} & 0.25 & 0.20 & 0.40 & 0.38 \\
    Kendall's $\tau$ & 1.02 & 0.86 & 1.43 & 1.41 & 0.03 & 0.02 \\
    LLM-as-Judge     & \textbf{1.38} & \textbf{0.21} & \textbf{1.16} & \textbf{0.24} & $-0.03$ & $-0.03$ \\
    \bottomrule
  \end{tabular}
\end{table}

\begin{table*}[t]
  \centering
  \small
  \setlength{\tabcolsep}{5pt}
  \renewcommand{\arraystretch}{1.1}
  \caption{Metric sensitivity $\Delta_m^{\mathrm{avg}}$ recomputed within
  workflow-size strata ($5$--$7$, $8$--$15$, and $16{+}$ nodes; $n{=}3842$,
  $1109$, and $22$ golden workflows). Sensitivities are broadly stable between
  the two well-populated strata. The $16{+}$ columns rest on only $22$
  workflows, so they are reported for completeness rather than as evidence about
  long industrial workflows; the largest divergences there (LLM-as-Judge under
  Missing Steps, and BERTScore and Kendall's $\tau$ under Compressed Steps) are
  shown in bold and are within the noise such a sample admits.}
  \label{tab:size-stratified}
  \begin{tabular}{@{}l ccc ccc ccc@{}}
    \toprule
    & \multicolumn{3}{c}{\textbf{Missing Steps}}
    & \multicolumn{3}{c}{\textbf{Compressed Steps}}
    & \multicolumn{3}{c}{\textbf{Description Changes}} \\
    \cmidrule(lr){2-4}\cmidrule(lr){5-7}\cmidrule(lr){8-10}
    \textbf{Metric} & 5--7 & 8--15 & 16+ & 5--7 & 8--15 & 16+ & 5--7 & 8--15 & 16+ \\
    \midrule
    Graph F1         & 0.73 & 0.70 & 0.69 & 1.04 & 1.03 & 0.81 & 0.30 & 0.25 & 0.38 \\
    Chain F1         & 0.73 & 0.70 & 0.78 & 1.04 & 1.03 & 0.84 & 0.30 & 0.26 & 0.41 \\
    BLEU             & 1.22 & 1.27 & 1.22 & 0.85 & 0.97 & 0.74 & 0.88 & 0.91 & 0.98 \\
    GLEU             & 0.95 & 1.00 & 0.96 & 0.81 & 0.93 & 0.70 & 0.83 & 0.87 & 0.93 \\
    BERTScore        & 0.20 & 0.35 & 0.32 & 0.20 & 0.31 & \textbf{0.68} & 0.39 & 0.38 & 0.33 \\
    Kendall's $\tau$ & 0.94 & 0.92 & 0.83 & 1.45 & 1.32 & \textbf{0.95} & 0.01 & 0.05 & 0.27 \\
    LLM-as-Judge     & 0.78 & 0.85 & \textbf{0.32} & 0.67 & 0.79 & 0.55 & $-0.03$ & $-0.01$ & $-0.02$ \\
    \bottomrule
  \end{tabular}
\end{table*}

\paragraph{Note on reproducing Table~\ref{tab:sensitivity-summary} from these
tables.} The sensitivity values $\Delta_m^{\mathrm{avg}}$ reported in
Table~\ref{tab:sensitivity-summary} are computed from the \emph{unrounded}
per-severity means, whereas the tables above display those means rounded to two
decimal places. Applying the formula
$\Delta_m^{\mathrm{avg}} = 2.5\cdot[\bar{s}_m(10\%) - \bar{s}_m(50\%)]$
to the rounded values in Tables~\ref{tab:score_change_missing}--\ref{tab:score_change_desc}
will therefore reproduce Table~\ref{tab:sensitivity-summary} only up to
$\pm 0.02$ per cell; this difference is rounding noise, not a discrepancy in
the underlying data.

\paragraph{Deriving the metric bundle and alert thresholds
(Table~\ref{tab:bundle-recommendation}).}
The Primary/Secondary assignments and operating-point thresholds in the
practical-guidance bundle are read directly off the sensitivity matrix
(Table~\ref{tab:sensitivity-summary}) and the per-severity score tables
(Tables~\ref{tab:score_change_missing}--\ref{tab:score_change_desc});
they are operating-point recommendations rather than tuned hyperparameters.

\emph{Primary/Secondary selection.} For each failure mode we take as
\textbf{Primary} the metric family that is both highly sensitive to that
perturbation \emph{and} diagnostically specific (comparatively unresponsive to
the unrelated modes), and as \textbf{Secondary} a complementary family that
covers the Primary's blind spot. (i)~\textsc{Missing Steps}: structural
Graph/Chain F1 ($\Delta=0.72$) localizes node loss, paired with LLM-as-Judge
($\Delta=0.80$) for semantic completeness; lexical metrics, although numerically
more sensitive ($\Delta_{\text{BLEU}}=1.23$), are not chosen as Primary because
they fire almost equally on all three modes ($0.88$ on both Compressed and
Description) and are therefore non-specific. (ii)~\textsc{Compressed Steps}:
Kendall's $\tau$ has the largest sensitivity ($\Delta=1.42$) and is near-inert
under Description changes ($0.02$), making it a specific ordering probe, paired
with structural F1 ($\Delta=1.04$). (iii)~\textsc{Description Changes}: lexical
BLEU/GLEU are the only family that moves while structure and order stay flat
(Kendall's $\tau$ $\Delta=0.02$, structural $0.29$), paired with BERTScore
($\Delta=0.39$) to bound whether paraphrasing altered meaning.

\emph{Caveat on specificity.} The specificity claims above are aggregate: they
are read from the mean sensitivities $\Delta_m^{\mathrm{avg}}$, which report
structural metrics as near-flat under \textsc{Description Changes} ($0.29$). This
inertness does not hold at the instance level: Graph F1 falls below $1.0$ for
$19.55\%$, $42.47\%$, and $59.18\%$ of paraphrase variants at $10/30/50\%$
(Section~\ref{sec:benchmark-construction}), because node alignment at
$\tau_{\text{align}}=0.8$ is itself sensitive to paraphrasing, so most such
declines do not indicate a change in topology. We therefore present the \textsc{Description Changes}
assignment---and the specificity argument supporting it---as an operating-point
recommendation at the current $\tau_{\text{align}}$, not as a derivation from
clean diagnostic separation. Re-deriving the bundle at a lower
$\tau_{\text{align}}$, where structural metrics are more nearly inert under
paraphrasing, is left to future work (see Limitations).

\emph{Threshold selection.} Each alert threshold is an operating point placed in
the separating band between the Primary metric's mean score under \emph{mild}
($10\%$) and \emph{moderate} ($30\%$) perturbation. Candidates
statistically consistent with $\le\!10\%$ degradation therefore pass, while those reaching
$\sim\!30\%$ degradation or worse are flagged. For \textsc{Missing Steps},
Graph/Chain F1 falls $0.90\!\to\!0.76\!\to\!0.61$ across $10/30/50\%$, so the
trip point is set at $0.75$ (the $30\%$ level). For \textsc{Description Changes},
BLEU/GLEU fall $\approx\!0.85\!\to\!0.66\!\to\!0.50$, giving a threshold of
$0.70$, just above the $30\%$ mean. For \textsc{Compressed Steps}, Kendall's
$\tau$ drops steeply and with high variance ($0.74\!\pm\!0.14\to0.46\!\pm\!0.16$);
the threshold is set more conservatively at $0.65$, between the $10\%$ and
$30\%$ means, to catch fast-onset ordering regressions earlier. These values
are defaults intended to be re-tuned to a deployment's own tolerance for missed
regressions versus false alerts.

\subsection{Instance-Level Gate Behavior}
\label{appendix:instance-gates}

Table~\ref{tab:bundle-recommendation}'s thresholds are derived from the \emph{mean}
score trajectories above, but a threshold placed at a severity's mean fires on only
about half the individual variants at that severity, because the score distribution
straddles it. Table~\ref{tab:instance-gates} reports the resulting instance-level
firing rates for each primary gate, and three consequences matter for CI/CD gating.
(i)~\emph{Missed regressions}: the \textsc{Missing Steps} gate (Graph/Chain
F1${<}0.75$) fires on only $58.4\%$ of $30\%$ deletions, so $41.6\%$ of genuine
$30\%$ deletions pass unflagged. (ii)~\emph{Mild-severity false alarms}: the
\textsc{Compressed Steps} gate (Kendall's $\tau{<}0.65$) fires on $21.0\%$ of $10\%$
compressions that the threshold design intends to pass.
(iii)~\emph{Benign-paraphrase false alarms}: the structural gate fires on $9.4\%$ and
$20.8\%$ of $30\%$ and $50\%$ \textsc{Description Changes}, whose target score is
$1.0$. These firings are driven in most cases by the SBERT-alignment effect of
Section~\ref{sec:benchmark-construction} rather than by topology changes. These are
less failures of the metrics than of reading a mean trajectory as an instance-level
guarantee; they reinforce that Table~\ref{tab:bundle-recommendation}'s thresholds are
operating points to be tuned to a deployment's tolerance for missed regressions
versus false alarms, ideally by choosing a point on each metric's
severity-conditioned score distribution rather than at its mean.

\begin{table*}[t]
  \centering
  \small
  \setlength{\tabcolsep}{5pt}
  \renewcommand{\arraystretch}{1.1}
  \caption{Instance-level firing rate of each Table~\ref{tab:bundle-recommendation}
  primary gate: the percentage of \emph{individual} variants whose primary-metric
  score trips the alert threshold, by perturbation type and severity. Bold marks
  each gate's own perturbation type (the mode it is the primary detector for); within
  that mode only the $30\%$ and $50\%$ columns are genuine regressions that should
  fire. \textsc{Description Changes} are benign (target score $1.0$), so \emph{any}
  firing there is a false alarm. A threshold placed at a severity's mean score fires
  on only about half the variants at that severity: e.g., the \textsc{Missing Steps}
  gate misses $41.6\%$ of the $30\%$ deletions it is meant to flag.}
  \label{tab:instance-gates}
  \begin{tabular}{@{}l ccc ccc ccc@{}}
    \toprule
    & \multicolumn{3}{c}{\textbf{Missing Steps}}
    & \multicolumn{3}{c}{\textbf{Compressed Steps}}
    & \multicolumn{3}{c}{\textbf{Description Changes}} \\
    \cmidrule(lr){2-4}\cmidrule(lr){5-7}\cmidrule(lr){8-10}
    \textbf{Primary gate} & 10\% & 30\% & 50\% & 10\% & 30\% & 50\% & 10\% & 30\% & 50\% \\
    \midrule
    Graph/Chain F1 $<0.75$ & \textbf{2.9} & \textbf{58.4} & \textbf{99.7}
                           & 15.0 & 77.9 & 99.9 & 1.6 & 9.8 & 21.5 \\
    Kendall's $\tau<0.65$  & 4.3 & 70.5 & 99.8
                           & \textbf{21.0} & \textbf{89.7} & \textbf{100.0} & 0.5 & 1.4 & 1.7 \\
    BLEU/GLEU $<0.70$      & 13.5 & 92.2 & 99.6 & 6.5 & 40.5 & 83.0
                           & \textbf{4.3} & \textbf{58.8} & \textbf{91.8} \\
    \bottomrule
  \end{tabular}
\end{table*}

\end{document}